%% file: ICASSP2024_CrowdandLLM.tex
\documentclass{article}
\usepackage{spconf,amsmath,graphicx}

\usepackage{amsmath}
\usepackage{multirow}
\usepackage{bm}
\usepackage{balance}
\usepackage{xurl}
\usepackage{bbm}
\usepackage{dsfont}
\usepackage{booktabs}

\usepackage{color}

\usepackage[numbers]{natbib}

\title{A Comparative Study on Annotation Quality of Crowdsourcing and LLM via Label Aggregation}
\name{Jiyi Li\thanks{This work was partially supported by JSPS KAKENHI Grant Number 23H03402 and JKA Foundation. }\thanks{Email:  garfieldpigljy@gmail.com}}
\address{University of Yamanashi, Kofu, Japan}
\begin{document}
%
\maketitle
\begin{abstract}
\input{contents/abstract}

\end{abstract}

\section{Introduction}
\input{contents/introduction}

\section{Label Aggregation}
\input{contents/problem}

\section{Empirical Study}
\input{contents/dataset}
\input{contents/experiments}

\section{Conclusions}
\input{contents/conclusion}

\input{contents/limitation}

\bibliographystyle{IEEEbib}
\bibliography{contents/crowd,contents/llm}

\end{document}

%% file: contents/abstract.tex
Whether Large Language Models (LLMs) can outperform crowdsourcing on the data annotation task is attracting interest recently. Some works verified this issue with the average performance of individual crowd workers and LLM workers on some specific NLP tasks by collecting new datasets. 
However, on the one hand, existing datasets for the studies of annotation quality in crowdsourcing are not yet utilized in such evaluations, which potentially provide reliable evaluations from a different viewpoint. 
On the other hand, the quality of these aggregated labels is crucial because, when utilizing crowdsourcing, the estimated labels aggregated from multiple crowd labels to the same instances are the eventually collected labels. 
Therefore, in this paper, we first investigate which existing crowdsourcing datasets can be used for a comparative study and create a benchmark. 
We then compare the quality between individual crowd labels and LLM labels and make the evaluations on the aggregated labels. In addition, we propose a Crowd-LLM hybrid label aggregation method and verify the performance. We find that adding LLM labels from good LLMs to existing crowdsourcing datasets can enhance the quality of the aggregated labels of the datasets, which is also higher than the quality of LLM labels themselves. 

%% file: contents/introduction.tex
Crowdsourcing has been widely used for data annotations. 
Because of the ability or diligence of the crowd workers, the quality of crowd labels is a crucial issue. 
Recently, Large Language Models (LLMs) have attracted interest from researchers because of their strong capability on diverse NLP tasks. One of the concerned issues is the capability on the data annotation tasks, i.e., whether LLMs can outperform crowdsourcing because LLMs are cheaper than crowd workers for labeling the instances. Veselovsky et al. \cite{veselovsky2023artificial} found that the crowd workers on MTurk have been recently using ChatGPT to complete the crowdsourcing tasks. 
Some works verified this issue with the average performance of individual crowd workers and LLM workers on some specific tasks by collecting new datasets for their target tasks \cite{CrowdChatgptSocialComputing,CrowdChatgptTextAnnotation,CrowdChatgptTwitter, CrowdChatgptIntentClassification}. In this paper, referring to the terms ``crowd worker" and ``crowd labels", we regard LLMs for data annotations as ``LLM workers" and denote the labels generated by LLMs as ``LLM labels". 

However, two issues are not yet well-studied. First, existing datasets for the studies of annotation quality in crowdsourcing are not yet utilized in such evaluations. In practice, without careful design on quality control during data collection, it may reach low-quality crowd data. Considering the volatile quality of crowdsourcing, such datasets can potentially provide reliable evaluations from a different viewpoint. 

A problem in utilizing the existing crowdsourcing datasets for verifying the LLM workers is that LLM workers need to access the text instance contents to assign the labels. 
Most existing datasets (e.g., \cite{venanzi2015weather, MultidimensionalWisdom}) do not contain the text instance contents. Therefore, in this work, we first investigate which existing crowdsourcing datasets can be used, create a benchmark, and make a comparative study on the quality between the individual crowd and LLM labels. 

Second, when utilizing crowdsourcing for data annotations, one type of methods for improving label quality is collecting multiple labels for each instance and estimating relatively high-quality labels from noisy crowd labels by label aggregation models. The aggregated labels are the eventually collected data, rather than the crowd labels themselves. The quality of these aggregated labels is thus imperative. Therefore, in this work, besides the comparisons of the data quality between individual crowd labels and LLM labels, we also make the evaluations on the aggregated labels on crowd labels or LLM labels. In addition, we propose a Crowd-LLM hybrid label aggregation method from both the individual crowd and LLM labels, and verify the performance. 

There are diverse types of labels in data annotations. 
For example, for the classification tasks, crowd workers are asked to provide categorical labels to the instances \cite{MultidimensionalWisdom,mj,ds,gladlong,cbcclong,hyperquestion,crowdworkersimilarity,crowdpr}. For some ranking tasks, workers are asked to provide pairwise preference labels, i.e., whether an instance is preferred to the other instance \cite{bt,crowdbtlong,HBTL,rankingclusteringlong,CollectivePreference}. 
There are also methods for other types of labels, such as numerical labels \cite{mj,catd} and textual labels \cite{textaggregation,ExplainNLPData}. In this paper, we mainly focus on the categorical labels. 

The main contributions of this work are as follows. 
    (1) We investigate which existing crowdsourcing datasets satisfy the purposes of this comparative study and propose a benchmark to provide reliable evaluations. 
    We compare the quality between individual crowd and LLM labels, which can also be regarded as the ability of individual crowd and LLM workers. 
    (2) We make the comparison from the viewpoint of estimated labels on crowd or LLM labels by label aggregation models. We propose a Crowd-LLM hybrid label aggregation on both crowd and LLM labels and verify the performance. 
    (3) We find that adding LLM labels to existing crowdsourcing datasets can enhance the quality of the aggregated labels of the datasets, which is also higher than the quality of LLM labels themselves. 

%% file: contents/problem.tex
We first formulate the problem setting of label aggregation for quality control in data annotation. 
We utilize both crowd and LLM workers for annotating categorical labels to instances. 
A typical task in crowdsourcing for collecting categorical labels is the multi-choice task with a single label. 
In such tasks, by showing the instances, the crowd and LLM workers are asked to select one label option from the given list of candidate choices for this instance, e.g., assigning a sentiment label (positive, neutral, or negative) to a tweet or selecting a choice from candidate answers to a quiz. 

The set of candidate labels for an instance is defined as $\mathcal{C}=\{c_k\}_{k=1}^\mathcal{K}$. We assume that $\mathcal{K}$ is the same for all instances, which is a typical setting. 
We define the set of workers as $\mathcal{A}=\{a_i\}_{i=1}^\mathcal{N}$, which can be crowd or LLM workers, and the set of instances as $\mathcal{B}=\{b_j\}_{j=1}^\mathcal{M}$. 
We define $y_{ij}$ as the label given by worker $a_i$ to instance $b_j$. We denote the set of all labels as $\mathcal{Y} = \{y_{ij}\}_{i,j}$, the set of labels given to $b_j$ as $\mathcal{Y}_{*j} = \{y_{ij} \vert a_i\in \mathcal{A} \}$ and the set of labels given by $a_i$ as $\mathcal{Y}_{i*} = \{y_{ij} \vert b_j\in \mathcal{B} \}$. 
Because the number of instances can be large, a worker is not necessary to annotate the entire set, while it is easier for a LLM worker to annotate the entire set than a crowd worker. 

For the setting of the label aggregation models, the inputs are worker set $\mathcal{A}$, instance set $\mathcal{B}$, and label set $\mathcal{Y}$. The outputs are the estimated true labels $\mathcal{\hat{Z}}=\{\hat{z}_j \}_j$. We use the common unsupervised setting in crowdsourcing that label aggregation models do not know the true labels because the true labels are unknown in the stage of data annotation. 

In traditional crowd label aggregation, the label set $\mathcal{Y}_{crowd}$ only consists of crowd labels.  When we regard LLMs as the workers and conduct LLM label aggregation, $\mathcal{Y}_{LLM}$ only consists of LLM labels. Intuitively, we can propose a \textit{Crowd-LLM hybrid label aggregation} method by collecting the hybrid label set $\mathcal{Y}_{hybrid}$ that consists of both individual crowd and LLM labels. The existing label aggregation models \cite{mj,ds,gladlong} proposed for crowd labels can be utilized on the hybrid label set $\mathcal{Y}_{hybrid}$ directly. 

%% file: contents/dataset.tex
\begin{table}[!t]
\scriptsize
\centering
\caption{Statistics of Datasets. $\vert\mathcal{Y}_{*j}\vert_{avg}$ ($\vert\mathcal{Y}_{i*}\vert_{avg}$): average number of labels per instance (per crowd worker).}
\label{tab:datasets}
\begin{tabular}{c|cccccc}
\toprule
Dataset & $\vert\mathcal{B}\vert$ & $\vert\mathcal{A}\vert$ & $\vert\mathcal{Y}\vert$ & $\mathcal{K}$ & $\vert\mathcal{Y}_{*j}\vert_{avg}$ & $\vert\mathcal{Y}_{i*}\vert_{avg}$ \\
\midrule
RTE & 800 & 164 & 8,000 & 2+1 & 10 & 48.78 \\
ENG & 30 & 63 & 1,890 & 5 & 63 & 30 \\
CHI & 24 & 50 & 1,200 & 5 & 50 & 24 \\
ITM & 25 & 36 & 900 & 4 & 36 & 25 \\
MED & 36 & 45 & 1,620 & 4 & 45 & 36 \\
POK & 20 & 55 & 1,100 & 6 & 55 & 20 \\
SCI & 20 & 111 & 2,220 & 5 & 110 & 20 \\
\bottomrule
\end{tabular}
\end{table}

\begin{table*}[!t]
\scriptsize
\centering
\caption{Accuracy of individual crowd workers and individual LLM workers. } 
\label{tab:single_worker_quality}
\begin{tabular}{c|cccc|ccc|ccc}
\toprule
\multirow{2}{*}{Dataset} & \multicolumn{4}{c|}{Crowd Worker} & \multicolumn{3}{c|}{ChatGPT} & \multicolumn{3}{c}{Vicuna} \\ 
& Min & Max & Mean & Median & $t$ = 0 & $t$ = 0.5 & $t$ = 1 & $t$ = 0 & $t$ = 0.5 & $t$ = 1 \\
\midrule
RTE & 0.400 & \textbf{1.00} & 0.837 & 0.850 & 0.859 & 0.849 & 0.821 & 0.739 & 0.739 & 0.739\\
ENG & 0.067 & \textbf{0.700} & 0.256 & 0.233 & 0.367 & 0.400 & 0.433 & 0.233 & 0.233 & 0.233 \\
CHI & 0.083 & 0.792 & 0.374 & 0.375 & \textbf{0.958} & \textbf{0.958} & 0.917 & 0.167 & 0.167 & 0.167 \\
ITM & 0.240 & 0.840 & 0.537 & 0.540 & 0.840 & \textbf{0.920} & 0.880 & 0.240 & 0.240 & 0.240 \\
MED & 0.194 & \textbf{0.917} & 0.475 & 0.417 & 0.750 & 0.750 & 0.778 & 0.222 & 0.222 & 0.222 \\
POK & 0.000 & \textbf{1.000} & 0.277 & 0.200 & 0.900 & 0.900 & 0.900 & 0.050 & 0.050 & 0.050 \\
SCI & 0.050 & \textbf{0.850} & 0.295 & 0.300 & 0.450 & 0.400 & 0.400 & 0.000 & 0.000 & 0.000 \\
\bottomrule
\end{tabular}
\end{table*}

\begin{table*}[!t]
\scriptsize
\setlength\tabcolsep{1pt}
\centering
\caption{Results of label aggregation and hybrid label aggregation. }
\label{tab:label_aggregation_results}
\begin{tabular}{c|ccc|ccc|ccc|ccc|ccc|ccc|ccc}
\toprule
Dataset & \multicolumn{3}{c|}{Crowd Worker Only} & \multicolumn{3}{c|}{ChatGPT Only} & \multicolumn{3}{c|}{Crowd + ChatGPT(1)} & \multicolumn{3}{c|}{Crowd + ChatGPT(3)} & \multicolumn{3}{c|}{Vicuna Only} & \multicolumn{3}{c|}{Crowd + Vicuna(1)} & \multicolumn{3}{c}{Crowd + Vicuna(3)} \\
\midrule
 & MV & DS & GLAD & MV & DS & GLAD & MV & DS & GLAD & MV & DS & GLAD & MV & DS & GLAD & MV & DS & GLAD & MV & DS & GLAD \\
\midrule
RTE & 0.919 & 0.801 & 0.928 & 0.850 & 0.606 & 0.850 & 0.920 & 0.815 & \textbf{0.935} & 0.911 & 0.775 & 0.928 & 0.739 & 0.739 & 0.739 & 0.900 & 0.824 & 0.918 & 0.856 & 0.674 & 0.859 \\
ENG & 0.467 & 0.400 & \textbf{0.533} & 0.400 & 0.300 & 0.400 & 0.433 & 0.400 & \textbf{0.533} & 0.400 & 0.400 & \textbf{0.533} & 0.233 & 0.167 & 0.233 & 0.433 & 0.367 & 0.500 & 0.433 & 0.367 & 0.467 \\
CHI & 0.625 & 0.542 & 0.583 & \textbf{0.958} & \textbf{0.958} & \textbf{0.958} & 0.625 & 0.625 & 0.667 & 0.708 & 0.708 & 0.792 & 0.167 & 0.167 & 0.167 & 0.625 & 0.542 & 0.583 & 0.625 & 0.542 & 0.583 \\
ITM & 0.760 & 0.680 & 0.720 & \textbf{0.880} & \textbf{0.880} & \textbf{0.880} & 0.760 & 0.800 & 0.760 & 0.840 & 0.840 & 0.840 & 0.250 & 0.250 & 0.250 & 0.720 & 0.720 & 0.720 & 0.720 & 0.720 & 0.720 \\
MED & 0.667 & 0.694 & 0.667 & 0.750 & 0.750 & 0.750 & 0.722 & 0.722 & 0.750 & 0.750 & 0.778 & \textbf{0.861} & 0.125 & 0.125 & 0.125 & 0.667 & 0.694 & 0.722 & 0.667 & 0.667 & 0.722 \\
POK & 0.650 & 0.600 & 0.850 & 0.900 & 0.900 & 0.900 & 0.700 & 0.650 & 0.900 & 0.800 & 0.800 & \textbf{1.000} & 0.000 & 0.000 & 0.000 & 0.650 & 0.600 & 0.850 & 0.650 & 0.600 & 0.850 \\
SCI & 0.550 & 0.550 & \textbf{0.600} & 0.450 & 0.500 & 0.450 & 0.550 & 0.550 & 0.550 & 0.550 & 0.550 & \textbf{0.600} & 0.000 & 0.000 & 0.000 & 0.550 & 0.550 & \textbf{0.600} & 0.550 & 0.550 & \textbf{0.600} \\
\bottomrule
\end{tabular}
\end{table*}

\begin{table*}[!t]
\scriptsize
\setlength\tabcolsep{2pt}
\centering
\caption{Results of label aggregation and hybrid label aggregation in the few-crowd setting. } 
\label{tab:few_crowd_label_aggregation_results}
\begin{tabular}{c|ccc|ccc|ccc}
\toprule
Dataset & \multicolumn{3}{c|}{Few-crowd Only}  & \multicolumn{3}{c|}{Few-crowd + ChatGPT(3)} & \multicolumn{3}{c}{Few-crowd + Vicuna(3)} \\
\midrule
 & MV & DS & GLAD & MV & DS & GLAD & MV & DS & GLAD \\
\midrule
RTE & 0.846 $\pm$ 0.022 & 0.730 $\pm$ 0.055 & 0.882 $\pm$ 0.020 & \textbf{0.897 $\pm$ 0.007} & 0.765 $\pm$ 0.009 & 0.878 $\pm$ 0.007 & 0.834 $\pm$ 0.014 & 0.566 $\pm$ 0.083 & 0.739 $\pm$ 0.001 \\
ENG & 0.290 $\pm$ 0.116 & 0.218 $\pm$ 0.084 & 0.282 $\pm$ 0.147 & \textbf{0.397 $\pm$ 0.054} & 0.376 $\pm$ 0.065 & \textbf{0.397 $\pm$ 0.014} & 0.255 $\pm$ 0.061 & 0.237 $\pm$ 0.083 & 0.234 $\pm$ 0.006 \\
CHI & 0.461 $\pm$ 0.114 & 0.375 $\pm$ 0.136 & 0.480 $\pm$ 0.140 & 0.897 $\pm$ 0.049 & 0.915 $\pm$ 0.075 & \textbf{0.957 $\pm$ 0.008} & 0.282 $\pm$ 0.078 & 0.310 $\pm$ 0.094 & 0.167 $\pm$ 0.004 \\
ITM & 0.676 $\pm$ 0.093 & 0.550 $\pm$ 0.133 & 0.683 $\pm$ 0.097 & 0.867 $\pm$ 0.043 & 0.866 $\pm$ 0.082 & \textbf{0.898 $\pm$ 0.024} & 0.533 $\pm$ 0.085 & 0.413 $\pm$ 0.120 & 0.310 $\pm$ 0.170 \\
MED & 0.569 $\pm$ 0.121 & 0.508 $\pm$ 0.137 & 0.603 $\pm$ 0.153 & 0.793 $\pm$ 0.048 & 0.790 $\pm$ 0.041 & \textbf{0.818 $\pm$ 0.032} & 0.431 $\pm$ 0.103 & 0.395 $\pm$ 0.143 & 0.303 $\pm$ 0.119 \\
POK & 0.366 $\pm$ 0.215 & 0.295 $\pm$ 0.206 & 0.380 $\pm$ 0.293 & 0.863 $\pm$ 0.085 & 0.768 $\pm$ 0.140 & \textbf{0.909 $\pm$ 0.025} & 0.345 $\pm$ 0.203 & 0.300 $\pm$ 0.211 & 0.353 $\pm$ 0.270 \\
SCI & 0.362 $\pm$ 0.131 & 0.306 $\pm$ 0.136 & 0.364 $\pm$ 0.148 & 0.447 $\pm$ 0.072 & 0.425 $\pm$ 0.083 & \textbf{0.452 $\pm$ 0.017} & 0.301 $\pm$ 0.126 & 0.272 $\pm$ 0.130 & 0.240 $\pm$ 0.110 \\
\bottomrule
\end{tabular}
\end{table*}

\subsection{Dataset Preparation}
For this study, we need the existing crowdsourcing datasets that have (1) text contents of instances, (2) individual crowd labels rather than only published aggregated labels, and (3) categorical labels. 
However, most of the NLP datasets have text contents but only provide the aggregated labels without individual crowd labels even though they are collected by crowdsourcing because they only use the aggregated labels for NLP tasks. 
On the other hand, the researchers major in the label quality and aggregation issues in the crowdsourcing area mainly focus on crowd labels regardless of the instance contents. A typical format of label data is a triplet (instance ID, worker ID, label). Some datasets do not provide any information on the text contents, e.g., the weather sentiment dataset \cite{venanzi2015weather}. 
Some datasets claim the resources of the original instances, but we find it challenging to match the instances between the crowdsourcing datasets and the original resources, e.g., the WSD dataset \cite{mj}. The text contents of some datasets were ever provided but are no longer available now, e.g., the sentiment judgment dataset provided at CrowdScale 2013 Shared Task challenge. 
Furthermore, some datasets provide instance contents with other media types,  such as images (e.g., \cite{MultidimensionalWisdom}). 
After investigating many datasets, we finally found several datasets that satisfied the requirements of our study. 

\noindent 
\textbf{RTE} (Recognizing Textual Entailment) \citep{mj}: It contains the binary labels of whether a sentence can be inferred from another sentence. Each crowd worker annotates a subset of instances. We allow LLM workers to output an ``unsure" label. For the other datasets in \citep{mj} that are not used in this work, the original text contents of the event temporal ordering dataset and word sense disambiguation dataset cannot be found or matched; the emotion dataset has numerical labels. 

\noindent
\textbf{QUIZ} \citep{hyperquestion}: It contains several datasets that a worker selects an answer from multiple candidates to a quiz on a specific topic, including: \textbf{ENG} (English), the most analogically similar word pair to a word pair; \textbf{CHI} (Chinese), the meaning of Chinese vocabularies; \textbf{ITM} (Information Technology), basic knowledge of information technology; \textbf{MED} (Medicine), about medicine efficacy and side effects; \textbf{POK} (Pokémon), the Japanese name of a Pok\'emon with English name; \textbf{SCI} (Science), intermediate knowledge of chemistry and physics. Each worker annotates all instances. 
The statistics of these datasets are listed in Table \ref{tab:datasets}. 
There is an issue that whether the LLMs have been trained on the datasets for evaluations. RTE is from common datasets in NLP and has this risk. The data contents of QUIZ are not from common datasets in NLP, they are almost unlikely ever used for training the LLMs.

\subsection{Evaluations with Individual crowd and LLM Labels}
We first evaluate the data quality between individual crowd workers and LLM workers. We utilize ChatGPT (GPT-3.5-turbo\footnote{Version: gpt-3.5-turbo-0613}) and Vicuna (Vicuna-13B\footnote{https://huggingface.co/vicuna/ggml-vicuna-13b-1.1/blob/main/ggml-old-vic13b-uncensored-q8\_0.bin}) \cite{vicuna2023}. For each LLM model, we utilize different temperature parameters $t$=\{0,0.5,1\}, and obtain three LLM workers. 
The outputs of LLMs with higher $t$ values have higher randomness. To facilitate the experiments, we did not generate multiple outputs for each $t$ and used multiple LLM workers with different $t$ to represent this randomness. We selected these two LLMs because ChatGPT is a typical commercial LLM that we can use through API with pricing plans, while Vicuna is a typical open-source model that people can deploy in the local servers for free. Typically, GPT-3.5-turbo has higher text generation ability\footnote{https://lmsys.org/blog/2023-06-22-leaderboard/} than Vicuna-13B. Thus, we regard ChatGPT as a \textit{good} LLM worker and Vicuna as a \textit{normal} LLM worker in the experiments and analysis. 

The contents and formats in the outputs of LLMs are relatively free; proper prompts are required to obtain the target labels. Besides, the rule-based post-processing on the outputs of LLMs is also required to fit the format of the outputs, which only contain the categorical labels. 
We use prompts and rules with minor differences for each LLM to generate the labels as well as possible. 
Furthermore, we let each LLM worker annotate all instances. 

The evaluation metric of label quality is accuracy, i.e., the percentage of correct labels in the labels of all instances. Table \ref{tab:single_worker_quality} shows the label quality of a single crowd or LLM worker. There are some observations on the datasets used in this study. \textit{The minimum accuracy of a crowd worker is very low, and the maximum accuracy of a crowd worker can be better than a good LLM worker (ChatGPT) in most cases. A good LLM worker is better than the mean and median of crowd workers.} \textit{Considering that in practical scenarios, we can utilize quality control methods such as qualification tests to only select high-ability crowd workers, crowd workers are still possibly better than good LLM workers.} Whether selecting crowd workers or good LLM workers depends on the detailed tasks. On the other hand, a normal LLM worker (Vicuna) is worse than the mean and median of crowd workers; although it is free, \textit{a normal LLM worker is still not good enough to replace the crowd workers}. 

%% file: contents/experiments.tex
\subsection{Evaluations with Label Aggregation}

Besides comparing the accuracy between the individual crowd and LLM workers, we also evaluate the quality of aggregation labels because it is the typical way when using crowdsourcing for data annotation. 
We utilize the following typical label aggregation models in this paper. 
(1) MV \cite{mj}: It is almost the simplest method, but in practical scenarios, people not majoring in crowdsourcing research usually use this method because it is easy to implement and the performance is not bad. 
(2) DS \cite{ds}: This is one typical method based on learning the confusion matrix of worker judgments. 
(3) GLAD \cite{gladlong}: This is one typical method of modeling worker ability and instance difficulty. 

We utilize the label aggregation models on the label sets of crowd workers (Table \ref{tab:label_aggregation_results} ``Crowd Worker Only") and the label sets of LLM workers (Table \ref{tab:label_aggregation_results} ``ChatGPT Only" and ``Vicuna Only"). 
Besides, for the proposed Crowd-LLM hybrid label aggregation on the hybrid label sets, 
for one LLM model, we merge the crowd label sets with the label sets of all three LLM workers (Table \ref{tab:label_aggregation_results} ``Crowd + ChatGPT(3)" and ``Crowd + Vicuna(3)"), or with the labels sets of the LLM worker with $t$=0 which can generate relatively stable outputs (Table \ref{tab:label_aggregation_results} ``Crowd + ChatGPT(1)" and ``Crowd + Vicuna(1)"). 

For a fair understanding of the results in Table \ref{tab:label_aggregation_results}, we compare the results for different (separate or hybrid) label sets with different worker settings by a specific label aggregation model. 
For example, in the columns of ``GLAD", ``Crowd Worker Only" and ``ChatGPT Only" outperform each other in all seven datasets; however, ``Crowd + ChatGPT(1)" outperforms or is equal to ``Crowd Worker Only" on six of seven datasets; ``Crowd + ChatGPT(3)" outperforms or is equal to ``Crowd Worker Only" on all seven datasets. In the columns of ``MV" or the columns of ``DS", there are similar observations. In addition, ``Crowd + ChatGPT(1)" and ``Crowd + ChatGPT(3)" actually also outperform or are equal to ``ChatGPT only" in five of seven datasets. 
Thus, we find that, \textit{especially for the existing crowdsourcing datasets, adding LLM labels from good LLMs can enhance the quality of the aggregated labels of the datasets, which is also higher than the quality of LLM labels themselves.} 
When creating a new dataset without any labels, considering the relatively high cost of crowd labels and the relatively low cost of LLMs, whether collecting the labels by crowd workers or good LLM workers depends on the tasks, and using quality control methods is recommended. 
Furthermore, \textit{the normal LLMs (Vicuna) cannot improve the quality of aggregation labels in most cases.}

Furthermore, as shown in Table \ref{tab:datasets}, the average number $\vert\mathcal{Y}_{*j}\vert_{avg}$ of labels for each instance is high. 
In practical scenarios, people always only collect a small number (e.g., five) of labels for each instance. We thus conduct another experiment by randomly sampling five labels for each instance in a dataset and creating subsets of crowd labels, named the ``Few-crowd" setting. We also merge the ``Few-crowd" labels with the LLM labels to create hybrid label sets. For each label set and a label aggregation model, we evaluate the average performance of label aggregation models on 100 trials. Table \ref{tab:few_crowd_label_aggregation_results} shows the results. \textit{The results of the ``Few-crowd" setting also support the observation that adding LLM labels from good LLMs can enhance the quality of aggregated labels.} 

There are also other observations. The results in Table \ref{tab:few_crowd_label_aggregation_results} with the Few-crowd setting are worse than that in Table \ref{tab:label_aggregation_results} (e.g., ``Crowd + ChatGPT(3)" in Table \ref{tab:label_aggregation_results} and ``Few-crowd + ChatGPT(3)" in Table \ref{tab:few_crowd_label_aggregation_results}), \textit{collecting more crowd labels is still possible to improve the quality aggregated labels, despite using both crowd and good LLM labels} (whether using the same budget to collect more LLM labels or more crowd labels is another issue). In both Table \ref{tab:label_aggregation_results} and \ref{tab:few_crowd_label_aggregation_results}, GLAD performs better than MV and DS, and using advanced aggregation models is recommended. 

%% file: contents/conclusion.tex
We investigated which existing crowdsourcing datasets are proper for the comparative study on the label quality between crowd workers and LLM workers. 
The re-organized datasets, crowd and LLM labels, label aggregation models, and the accuracy of aggregated labels form a benchmark that can be used for a reliable comparative study. 
We compare the quality between individual crowd labels and LLM labels and make the evaluations on the aggregated labels. The comparisons based on the proposed Crowd-LLM hybrid label aggregation method show that adding good LLM labels to existing crowdsourcing datasets can enhance the quality of aggregated labels, which is also higher than the quality of LLM labels. 
Other observations include, e.g., (1) if using quality control methods such as qualification test so that we can only
select high-ability crowd workers for annotation, crowd workers are still possibly better than good LLM workers (e.g., ChatGPT); (2) normal LLM workers (e.g., Vicuna) still cannot replace the crowd workers; (3) collecting more crowd labels is still possible to improve the quality of aggregated labels even though using both crowd and good LLM labels. 

%% file: contents/limitation.tex
The limitation is that this work has only covered categorical labels. 
There are still other types of labels (e.g., numerical labels \cite{catd}, pairwise comparison labels \cite{rankingclusteringlong,CollectivePreference}, triplet comparison labels \cite{crowdtriplet,multiviewcrowdtriplet}, and text sequence labels \cite{textaggregationdataset,textaggregation,ExplainNLPData}) that are not yet included. 
We will continue investigating the datasets and comparative studies for other types of labels. 

%% file: ICASSP2024_CrowdandLLM.bbl
\begin{thebibliography}{10}

\bibitem{veselovsky2023artificial}
Veniamin Veselovsky, Manoel~Horta Ribeiro, and Robert West,
\newblock ``Artificial artificial artificial intelligence: Crowd workers widely use large language models for text production tasks,''
\newblock in {\em arXiv}, 2023.

\bibitem{CrowdChatgptSocialComputing}
Yiming Zhu, Peixian Zhang, Ehsan-Ul Haq, Pan Hui, and Gareth Tyson,
\newblock ``Can chatgpt reproduce human-generated labels? a study of social computing tasks,''
\newblock in {\em arXiv}, 2023.

\bibitem{CrowdChatgptTextAnnotation}
Fabrizio Gilardi, Meysam Alizadeh, and Maël Kubli,
\newblock ``Chatgpt outperforms crowd-workers for text-annotation tasks,''
\newblock in {\em arXiv}, 2023.

\bibitem{CrowdChatgptTwitter}
Petter Törnberg,
\newblock ``Chatgpt-4 outperforms experts and crowd workers in annotating political twitter messages with zero-shot learning,''
\newblock in {\em arXiv}, 2023.

\bibitem{CrowdChatgptIntentClassification}
Jan Cegin, Jakub Simko, and Peter Brusilovsky,
\newblock ``Chatgpt to replace crowdsourcing of paraphrases for intent classification: Higher diversity and comparable model robustness,''
\newblock in {\em arXiv}, 2023.

\bibitem{venanzi2015weather}
Matteo Venanzi, WTL Teacy, Alex Rogers, and Nicholas~R Jennings,
\newblock ``Weather sentiment - amazon mechanical turk dataset,''
\newblock 2015.

\bibitem{MultidimensionalWisdom}
Peter Welinder, Steve Branson, Serge Belongie, and Pietro Perona,
\newblock ``The multidimensional wisdom of crowds,''
\newblock in {\em NIPS}, 2010, pp. 2424--2432.

\bibitem{mj}
R.~Snow, B.~O'Connor, D.~Jurafsky, and A.~Y. Ng,
\newblock ``Cheap and fast---but is it good?: Evaluating non-expert annotations for natural language tasks,''
\newblock in {\em EMNLP}, 2008, pp. 254--263.

\bibitem{ds}
A.~P. Dawid and A.~M. Skene,
\newblock ``Maximum likelihood estimation of observer error-rates using the em algorithm,''
\newblock {\em Applied Statistics}, vol. 28, no. 1, pp. 20--28, 1979.

\bibitem{gladlong}
Jacob Whitehill, Paul Ruvolo, Tingfan Wu, Jacob Bergsma, and Javier Movellan,
\newblock ``Whose vote should count more: Optimal integration of labels from labelers of unknown expertise,''
\newblock in {\em NIPS}, 2009, pp. 2035--2043.

\bibitem{cbcclong}
Matteo Venanzi, John Guiver, Gabriella Kazai, Pushmeet Kohli, and Milad Shokouhi,
\newblock ``Community-based bayesian aggregation models for crowdsourcing,''
\newblock in {\em WWW}, 2014, pp. 155--164.

\bibitem{hyperquestion}
Jiyi Li, Yukino Baba, and Hisashi Kashima,
\newblock ``Hyper questions: Unsupervised targeting of a few experts in crowdsourcing,''
\newblock in {\em CIKM}, 2017, pp. 1069--1078.

\bibitem{crowdworkersimilarity}
Jiyi Li, Yukino Baba, and Hisashi Kashima,
\newblock ``Incorporating worker similarity for label aggregation in crowdsourcing,''
\newblock in {\em ICANN}, 2018, pp. 596--606.

\bibitem{crowdpr}
Jiyi Li, Yasushi Kawase, Yukino Baba, and Hisashi Kashima,
\newblock ``Performance as a constraint: An improved wisdom of crowds using performance regularization,''
\newblock in {\em IJCAI}, 2020, pp. 1534--1541.

\bibitem{bt}
Ralph~Allan Bradley and Milton~E Terry,
\newblock ``Rank analysis of incomplete block designs: I. the method of paired comparisons,''
\newblock {\em Biometrika}, vol. 39, no. 3/4, pp. 324--345, 1952.

\bibitem{crowdbtlong}
Xi~Chen, Paul~N. Bennett, Kevyn Collins-Thompson, and Eric Horvitz,
\newblock ``Pairwise ranking aggregation in a crowdsourced setting,''
\newblock in {\em WSDM}, 2013, pp. 193--202.

\bibitem{HBTL}
Tao Jin, Pan Xu, Quanquan Gu, and Farzad Farnoud,
\newblock ``Rank aggregation via heterogeneous thurstone preference models,''
\newblock in {\em AAAI}, 2020, vol.~34, pp. 4353--4360.

\bibitem{rankingclusteringlong}
Jiyi Li, Yukino Baba, and Hisashi Kashima,
\newblock ``Simultaneous clustering and ranking from pairwise comparisons,''
\newblock in {\em IJCAI}. 7 2018, pp. 1554--1560, International Joint Conferences on Artificial Intelligence Organization.

\bibitem{CollectivePreference}
Jiyi Li,
\newblock ``Context-based collective preference aggregation for prioritizing crowd opinions in social decision-making,''
\newblock in {\em WWW}, 2022, pp. 2657--2667.

\bibitem{catd}
Qi~Li, Yaliang Li, Jing Gao, Lu~Su, Bo~Zhao, Murat Demirbas, Wei Fan, and Jiawei Han,
\newblock ``A confidence-aware approach for truth discovery on long-tail data,''
\newblock {\em Proc. VLDB Endow.}, vol. 8, no. 4, pp. 425–436, 2014.

\bibitem{textaggregation}
Jiyi Li,
\newblock ``Crowdsourced text sequence aggregation based on hybrid reliability and representation,''
\newblock in {\em SIGIR}, 2020, pp. 1761--1764.

\bibitem{ExplainNLPData}
Sarah Wiegreffe and Ana Marasovic,
\newblock ``Teach me to explain: A review of datasets for explainable natural language processing,''
\newblock in {\em Proceedings of the Neural Information Processing Systems Track on Datasets and Benchmarks}, 2021, vol.~1.

\bibitem{vicuna2023}
Wei-Lin Chiang, Zhuohan Li, Zi~Lin, and et~al.,
\newblock ``Vicuna: An open-source chatbot impressing gpt-4 with 90\%* chatgpt quality,''
\newblock March 2023.

\bibitem{crowdtriplet}
Jiyi Li, Lucas~Ryo Endo, and Hisashi Kashima,
\newblock ``Label aggregation for crowdsourced triplet similarity comparisons,''
\newblock in {\em ICONIP}, 2021, pp. 176--185.

\bibitem{multiviewcrowdtriplet}
Xiaotian Lu, Jiyi Li, Koh Takeuchi, and Hisashi Kashima,
\newblock ``Multiview representation learning from crowdsourced triplet comparisons,''
\newblock in {\em WWW}, 2023, p. 3827–3836.

\bibitem{textaggregationdataset}
Jiyi Li and Fumiyo Fukumoto,
\newblock ``A dataset of crowdsourced word sequences: Collections and answer aggregation for ground truth creation,''
\newblock in {\em AnnoNLP2019}, Nov. 2019, pp. 24--28.

\end{thebibliography}
